\documentclass{article} 
\usepackage{iclr2021_conference,times}

\usepackage{amsmath,amsfonts,bm}

\def\eqref#1{equation~\ref{#1}}

\def\1{\bm{1}}

\DeclareMathAlphabet{\mathsfit}{\encodingdefault}{\sfdefault}{m}{sl}
\SetMathAlphabet{\mathsfit}{bold}{\encodingdefault}{\sfdefault}{bx}{n}

\usepackage{hyperref}
\usepackage{url}
\usepackage{graphicx}
\usepackage{svg}
\usepackage{tabularx}
\usepackage{booktabs}
\usepackage{placeins}
\usepackage{enumitem}
\usepackage{xcolor}    
\title{Wan-S2V: Audio-Driven Cinematic Video Generation}

\author{HumanAIGC Team\\
Tongyi Lab, Alibaba}

\iclrfinalcopy 
\makeatletter
\renewcommand{\headrulewidth}{0pt} 
\begin{document}

\maketitle

\begin{abstract}
Current state-of-the-art (SOTA) methods for audio-driven character animation demonstrate promising performance for scenarios primarily involving speech and singing. However, they often fall short in more complex film and television productions, which demand sophisticated elements such as nuanced character interactions, realistic body movements, and dynamic camera work.
To address this long-standing challenge of achieving film-level character animation, we propose an audio-driven model, which we refere to as Wan-S2V, built upon Wan. Our model achieves significantly enhanced expressiveness and fidelity in cinematic contexts compared to existing approaches.

We conducted extensive experiments, benchmarking our method against cutting-edge models such as Hunyuan-Avatar and Omnihuman. The experimental results consistently demonstrate that our approach significantly outperforms these existing solutions. Additionally, we explore the versatility of our method through its applications in long-form video generation and precise video lip-sync editing.
\end{abstract}

\section{Introduction}
\label{intro}
Audio-driven human video generation has made significant progress recently, largely thanks to the development and application of Diffusion models~\cite{ddpm}. Beginning with UNet-based text-to-image models and progressing to the latest DiT-based text-to-video models~\cite{wan2025,hunyuanvideo}, the quality of visual generation has also dramatically improved. Consequently, audio-driven models leveraging these latest DiT-based video foundation models are garnering increasing research attention~\cite{ominihuman,hu2025HunyuanVideo-Avatar,Wang2025FantasyTalkingRT}. However, influenced by prior work, current research primarily confines audio-driven models to single-scene human video generation, or even solely to single-character video driving. Nevertheless, in more complex scenarios such as film and television productions or multi-person scenes, audio-driven models still face numerous challenges. For instance, key questions arise: How can audio control a character while ensuring their movements are consistent and coordinated with the overall scene? How can person interactions be managed by audio and prompt jointly?
This paper primarily focuses on audio-driven human generation in such complex scenarios as film and television, aiming to enhance the efficacy of audio-driven generation through comprehensive data acquisition, robust model training, and clever yet effective inference strategies.

Achieving film-quality audio-driven video, we contend, requires simultaneously leveraging the distinct yet complementary capabilities of text and audio. From a practical user perspective, text is optimally utilized for delineating the overarching dynamics of the video, including cinematic camera movements, comprehensive character trajectories, and interactions between entities. Audio, conversely, excels at dictating minute details such as character expressions and localized actions, including precise hand gestures and head orientation.
Firstly, we construct our audio-driven model by leveraging the latest Wan text-to-video foundation model\cite{wan2025}. Our aim is to integrate audio-driven capabilities while preserving its inherent text control. Crucially, to ensure our model maintains text-control fidelity during training, we utilized Qwen-VL's~\cite{Qwen2.5-VL} video understanding capabilities for detailed textual captioning of videos, with a particular emphasis on descriptions pertinent to character motion.
To effectively support generation in complex scenarios, such as film and television productions, we curated film and television-related audio-visual data from existing open-source datasets and augmented it with our own internally collected dataset of talking and singing character videos to form our comprehensive training dataset.
While some existing methods attempt to reduce training complexity by training only partial network parameters, this often leads to conflicts between text and audio control. We hypothesize that a larger model capacity is more conducive to learning superior and harmonious text and audio control, thereby mitigating such conflicts. To facilitate large-scale, full-parameter training, and drawing inspiration from established parallel training paradigms for video foundation models, we implemented a hybrid training strategy combining FSDP~\cite{fsdp} with Context Parallel, significantly accelerating the training process.
Furthermore, to ensure enhanced stability and performance, we employed a multi-stage training regimen. This includes pre-training of the audio processing modules, followed by a comprehensive pre-training phase on the entire dataset, and subsequent fine-tuning on high-quality data. Collectively, these systematic strategies enable us to develop a robust and efficient audio-driven human video generation model.

Long video generation is crucial for generating videos in film and television scenarios. However, it faces challenges in maintaining stable details and consistency in scenes and even motion. Audio-driven methods like~\cite{tian2024emo} have attempted to use Motion Frames to maintain consistency between multiple clips, but an excessive number of motion frames can drastically increase computational complexity. This leads to a relatively limited number of motion frames, making it difficult to maintain long-term video stability in film and television scenarios. To address this, we introduce a approach similar to ~\cite{zhang2025framepack}, which effectively reduces the token count of Motion Frames by employing different token compression ratios at different times. This ensures the incorporation of more Motion Frames, thereby enabling the generation of more stable long videos.


To train our model, we constructed a dataset containing over clips, based on both publicly available video datasets and our own collected video data. This comprehensive dataset includes videos from solo scenarios focusing on human speech and singing, as well as complex character videos from film and television dramas. 

Our main contributions are as follows:
\begin{itemize}
    \item \textbf{Extending Audio-Driven Generation to Complex Scenarios:} We go beyond talking heads by enabling the creation of natural and expressive character movements in diverse and challenging scenes, incorporating both text-guided global motion control and audio-driven fine-grained local motion.
    \item \textbf{Long Video Stabilization and Efficient Model Variants:} We tackle the challenges of long video generation through optimized motion frame token reduction.
    \item \textbf{Comprehensive Training Data}We leverage a large-scale, diverse dataset to train our model and validate the effectiveness of our model through extensive experiments.
\end{itemize}

\section{Data Processing Pipeline}

\textbf{Data Collection.} Human-driven narratives constitute the core element of video content. Our objective is to identify videos featuring one or more human characters engaged in specific activities. Specifically, we adopted a two-pronged strategy:

\textit{Automated screening of large-scale datasets.} We collected videos from open source video datasets such as \cite{OpenHumanVid} and \cite{koala36m}, followed by an initial coarse filtering process that detected the presence of human-related descriptions in video captions. It is worth noting that the captions provided by these datasets are inherently coarse-grained and often fail to capture the nuanced, dynamic activities performed by characters (e.g., complex gestures, interactions, or context-specific behaviors). To address this limitation, we developed a specialized captioning pipeline designed to focus on human motion patterns, which will be elaborated in subsequent subsections.

\textit{Manual curation of high-quality samples.} Complementing the above approach, we manually selected videos containing intentional and complex human activities (e.g. speaking, singing, dancing) from public accessible sources. This dual methodology yielded an initial video pool comprising millions of human-centric video samples, forming the foundation for our dataset. 
\begin{figure*}[tb]
  \centering
  \includegraphics[width=1.0\textwidth]{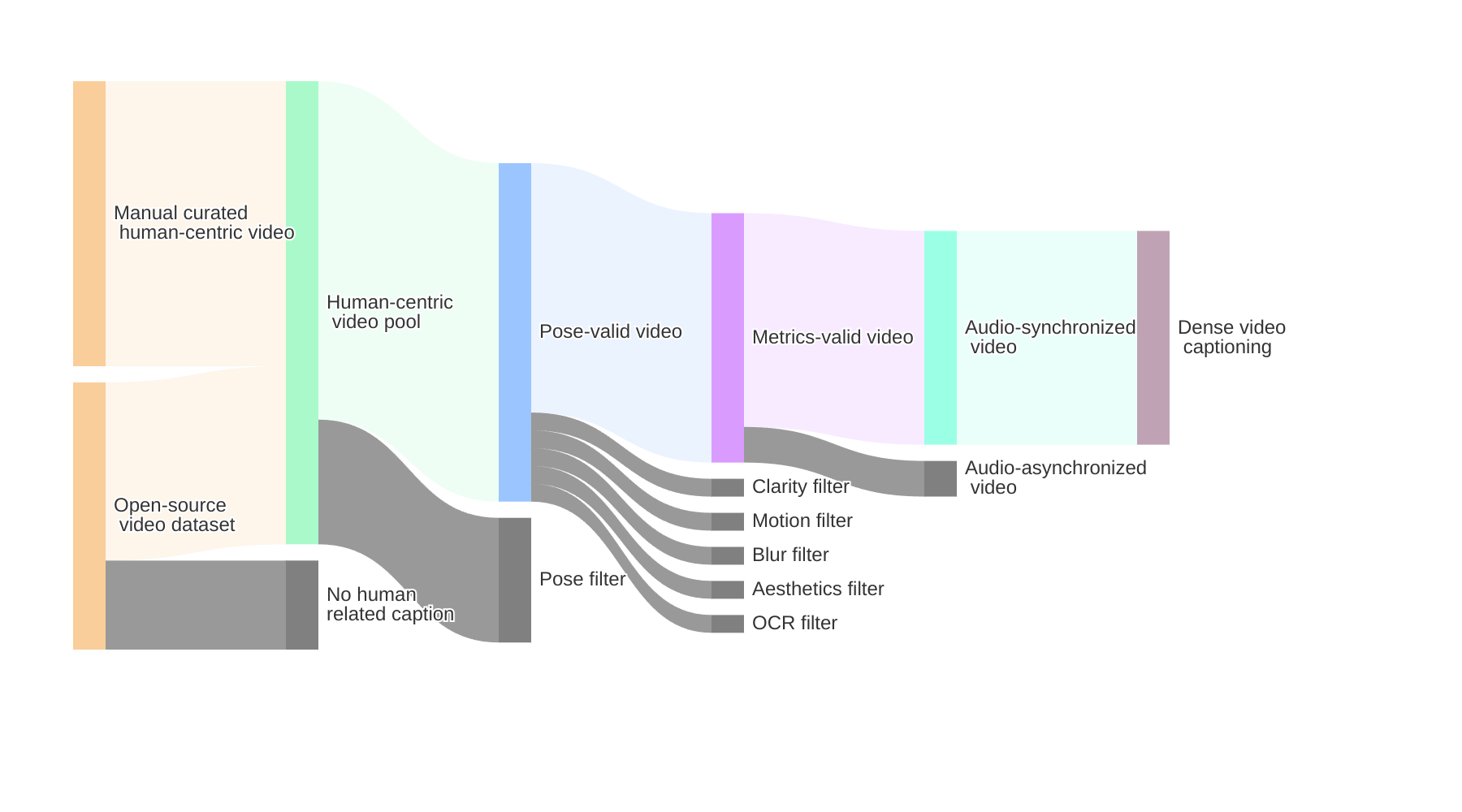}
  \caption{Overview of our hierarchical human-centric video filtering pipeline.
  }
  \label{fig:data_filtering_pipeline}
\end{figure*}

\textbf{Pose Tracking and Fine-grained Filtering.}  From the initial human-centric video pool, the 2D pose of each character is tracked via VitPose \cite{vitpose} and converted to DWPose \cite{dwpose}. This pose information serves two critical functions: (1) As a multi-modal control signal: The tracked pose is integrated as an optional multi-modal control signals for our human-centric video generation model, enabling precise temporal alignment with human actions. (2) For dataset refinement: The pose data is further leveraged to implement a fine-grained screening process. Specifically, we filtered out videos where characters occupy only a negligible portion in either temporal or spatial dimensions. Additionally, to ensure the model can learn audio-driven facial expressions from the given audio signals, we retained only videos containing consistent and visible human faces throughout the sequence.
Complementing the pose-based screening, we employed pre-trained video quality assessment models to evaluate motion extent, aesthetic appeal, and visual clarity. Videos were subsequently filtered based on these quantitative metrics to maintain high data quality. Furthermore, to address audio-visual alignment challenges, we utilized Light-ASD \cite{light-asd} to detect and exclude videos where (1) the audio is not synchronized with the active speaker, or (2) no active speaker exists in the scene.

\textbf{Video Quality.} To comprehensively evaluate video quality from multiple perspectives, we employ the following five metrics: (1) Clarity Assessment: We utilize the Dover metric \cite{dover} to quantify video clarity, which measures the perceptual sharpness of visual content. (2) Motion Stability Analysis: To evaluate temporal coherence, we predict optical flow using the UniMatch framework \cite{unimatch} and calculate a motion score. This helps identify and filter videos with excessive subject/background movement that could compromise visual quality. (3) Facial/Hand Sharpness Verification: A Laplacian operator is applied specifically to human faces and hands within the video frames. This technique enables the detection and exclusion of videos containing blurred facial features or hand regions. (4) Aesthetic Quality Evaluation: We incorporate an improved aesthetic predictor \cite{improved-aesthetic-predictor} to assess visual appeal based on human aesthetic preferences, ensuring the output meets subjective quality standards. (5) Subtitle Occlusion Detection: An OCR-based detector is applied to identify and exclude cases where subtitles might occlude faces or hands in video.

\textbf{Dense Video Caption.} A detailed and accurate video caption facilitates the alignment of the generation model with the input prompt. We used QwenVL2.5-72B \cite{Qwen2.5-VL} to generate captions for the videos, instructing the model to describe the following key aspects in details: (1) Camera angles, such as straight-on, overhead, low-angle, wide shot, medium shot, and close-up; (2) Physical appearance features (e.g., clothing and accessories) and actions, broken down into specific movements of the subject; (3) Main features of the background environment, including architectural style, color schemes, and greenery, among others.
At the same time, we required the model to avoid subjective evaluations and emotional interpretations, which are often trivial to generating the expected video content

\section{Model Architecture}

\begin{figure*}[tb]
  \centering
  \includegraphics[width=1.0\textwidth]{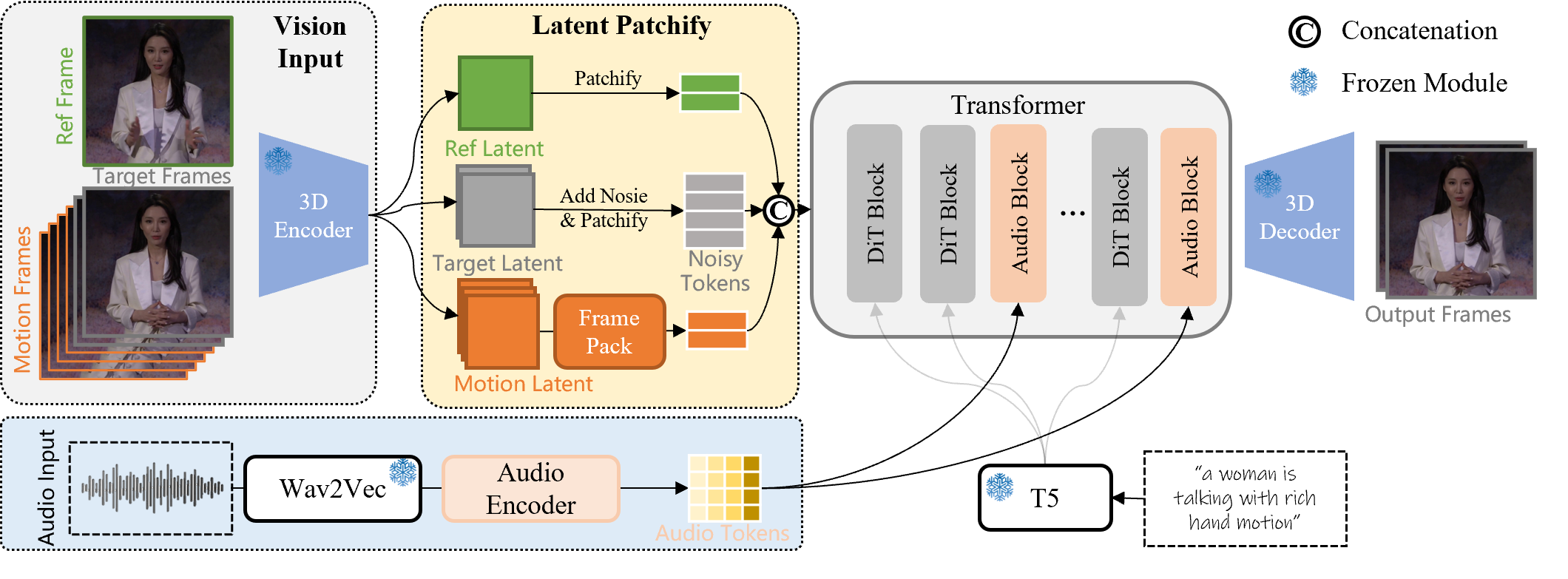}
  \caption{Overview of our pipeline.
  }
  \label{fig:pipeline1}
\end{figure*}

Given a single reference image, an input audio and a prompt to describe the video content, we could generate the video synchronized with the audio while preserving the content in the reference image (not start from the image).
As shown in Fig~\ref{fig:pipeline1}, our work is fed with multi-frame noise latent input, and tries to denoise them to the consecutive video frames during each timestep.

During the training, the RGB target frames $X\in\mathbb{R}^{F\times{H}\times{W}\times{3}}$ are encoded by 3D VAE into latent presentation $x_0\in\mathbb{R}^{f\times{h}\times{w}\times{c}}$, assigning a continuous time step $t\in[0,1]$, the noise $\epsilon$ are added to $x_0$ to get noisy latent $x_t$ according to Flow Matching introduced by \cite{lipman2023flowmatchinggenerativemodeling}:
$$ x_t=t\epsilon+(1-t)x_0$$
Input the noisy representation $x_t$, the target of the model is to predict the velocity $\frac{dx}{dt}=\epsilon-x_0$. During the inference, the model recoveres the noisy input $x_t$ into $x_0$ under the condition of the reference frame, motion frames, audio input and prompt.

The reference image, the target frames and the motion frames following \cite{tian2024emo} are fed into 3D VAE to down-sample the video spatially and temporally, getting the latent representation of the frames. All latent frames are then patchified and flattened, they are concatenated to be a sequence of visual tokens. The motion frames are optional, they provide the condition of the previous information, making the generated clips continuous. In order to generate long-term consistent video frames, it is necessary to obtain more historical information, since directly flatten the motion latent token could introduce more computational load. The motion latent is further compressed by Frame Pack module introduced by \cite{zhang2025framepack}, which compress the earlier frames in higher compressibility.

\begin{figure*}[tb]
  \centering
  \includegraphics[width=0.6\textwidth]{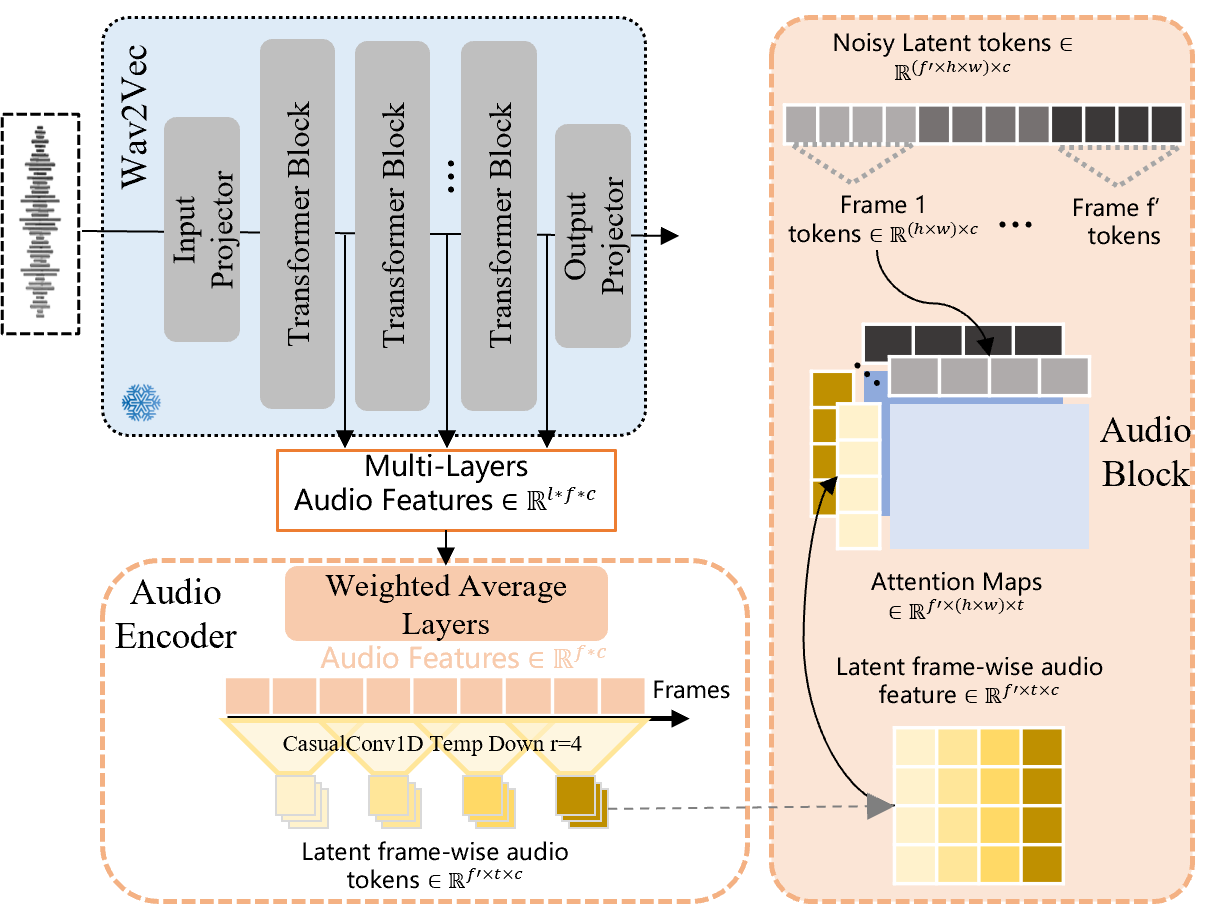}
  \caption{The pipeline of the audio injection.
  }
  \label{fig:audio_pipe}
\end{figure*}

As illustrated in Figure~\ref{fig:audio_pipe}, the raw audio waveform is first encoded using Wav2Vec by \cite{wav2vec}. To comprehensively capture the audio features, we adopt the weighted average layer proposed by \cite{tian2024emo}, which combines features from different layers through learnable weights. This approach effectively integrates shallow-level rhythmic and emotional cues with deep-level lexical content features extracted by Wav2Vec, thereby enhancing synchronization with complex audio signals such as singing or expressive speech. The resulting frame-wise audio features are then compressed along the temporal dimension using multiple causal 1D convolutional modules. This process generates audio features of the $i$ th latent frame $a_i\in\mathbb{R}^{f\times{t}\times{c}}$ that are temporally aligned with the video latent frames, where $t$ denotes the number of audio tokens per latent frame.

The latent audio features $a$ are passed into each Audio Block, where the noisy latent tokens $x_t\in\mathbb{R}^{(f'\times{h}\times{w})\times{c}}$ are divided into segments $\sum_i^{f'}x_{ti}\in\mathbb{R}^{(h\times{w})\times{c}}$ along the temporal dimension. 
To reduce computational overhead, attention is calculated between $a_i$ and $x_{ti}$, rather than performing full 3D attention between visual tokens and audio tokens. This approach ensures that the audio features and visual tokens are naturally synchronized.

\section{Implementation}
To train our Audio-to-Video model, we adopted a hybrid-parallel training scheme which combines FSDP and context parallelism, enabling large-scale, full-parameter model training. To support different resolutions, we support the training of variable-length video data. Our model is trained based on a pre-trained Wan model and is designed with a three-stage training process, including: audio encoder training, training on speech videos, training on film and television + speech videos, and finally, high-quality SFT (Supervised Fine-Tuning) stage training.
\subsection{Parallel Strategy}
To efficiently train our large model, a hybrid parallel training strategy is employed. This involves combining Fully Sharded Data Parallelism (FSDP)~\cite{fsdp} with Context Parallelism. Initially, FSDP is leveraged to shard the model's parameters across 8 GPU cards within a single node, enabling the training of our Wan-S2V-14B model while utilizing 80GB of memory per GPU.

Subsequently, for parallel computing, we implement a Context Parallelism scheme, combining RingAttention and Ulysses similar to \cite{usp}. This integrated approach, executed on 8 GPUs within a single node, allows us to achieve near-linear speedup, significantly reducing the single training iteration time from $\sim$100 seconds to $\sim$12 seconds. This robust setup ultimately supports the training of models exceeding 16B parameters, including our audio encoder and cross-attention components, enabling high-resolution video training up to 48 frames at $1024\times 768$ resolution (Height $\times$ Width) on 8 GPUs.

To accommodate diverse output resolutions and optimize training, a variable-length resolution training method is implemented. This method uses the token count, determined after the patchify operation, as a key metric. A maximum allowable token limit, $M$, is established. For videos exceeding this limit, resolution resizing or cropping is applied to reduce the token count to $M$ or below. Videos with token counts already below $M$ are used directly for model training without any modifications.


\section{Experiments}
Following the data construction pipeline detailed in Section 3, we meticulously filtered data from the OpenHumanViD~\cite{OpenHumanVid} dataset and integrated it with our self-constructed internal talking head dataset to form our comprehensive training set.

We constructed the audio-driven human video generation model on Wan-14B referred to as Wan-S2V-14B.

In comprehensive comparisons against existing state-of-the-art audio-driven video generation models, both quantitative metrics and visual results consistently demonstrate that our method surpasses current approaches in terms of expressiveness and the realism of generated content.

\subsection{Qualitative Evaluation}

\textbf{Comparison with SOTA}

\begin{figure*}[htbp]
\centering
\includegraphics[scale=0.16]{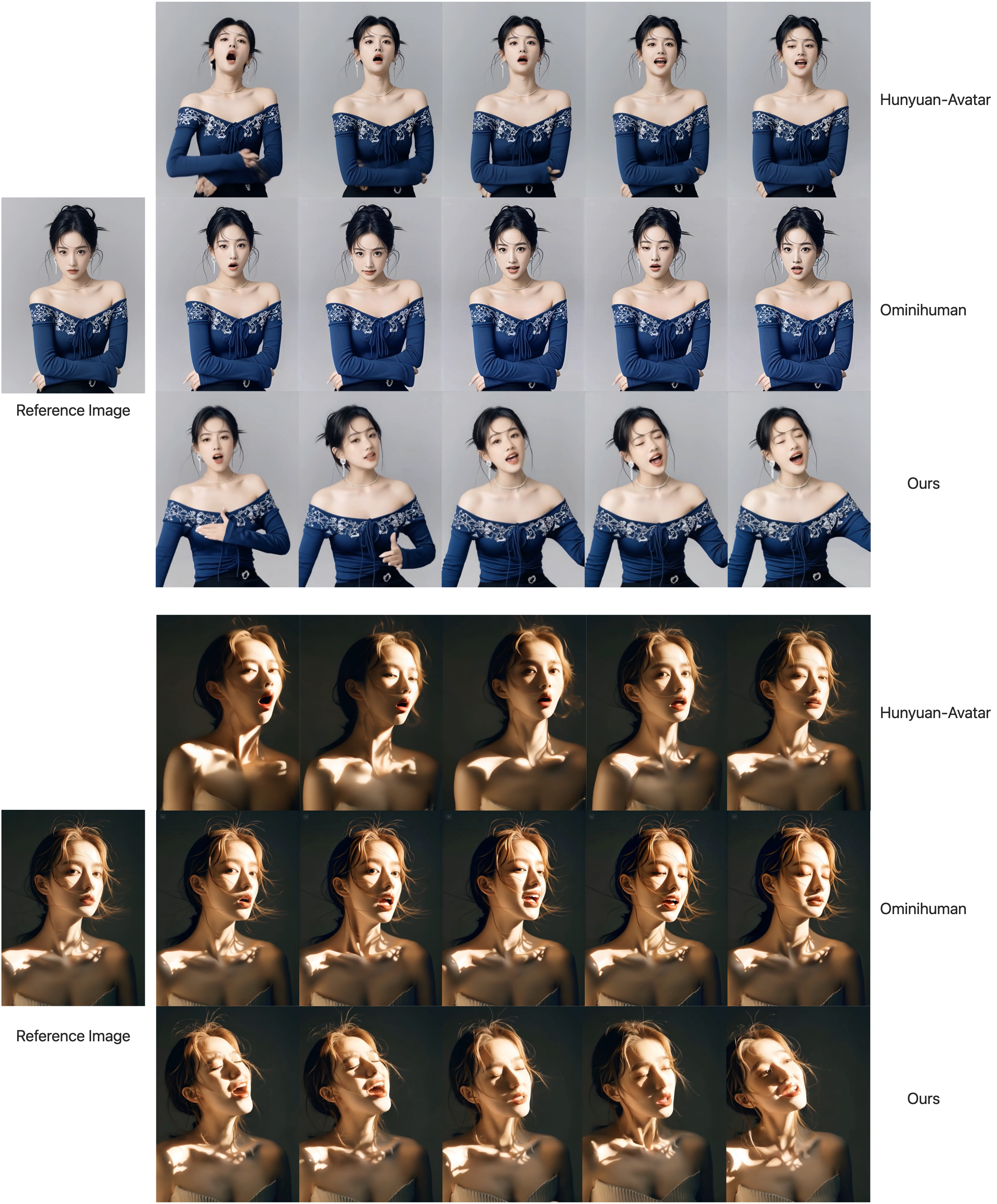}
\caption{Qualitative comparison of generated human videos. The leftmost column displays the reference image. Hunyuan-Avatar (top row) often suffers from facial distortions and inconsistent identity during large movements. Ominihuman (middle row) typically generates results with a limited range of motion, largely adhering to the pose of the reference image. In contrast, our method (bottom row) achieves superior performance in both motion dynamics and identity consistency.}
\label{fig:comparison}
\end{figure*}

A comparative study was conducted between our method and two existing DiT-based audio-driven video generation models, Ominihuman proposed by \cite{ominihuman} and Hunyuan-Avatar proposed by ~\cite{hu2025HunyuanVideo-Avatar}, revealing the superior capabilities of our approach. Figure~\ref{fig:comparison} illustrates these findings: Hunyuan-Avatar struggles with facial distortion and identity shifts during large-scale movements, while our model excels at maintaining identity consistency even amidst highly dynamic motion. Additionally, Ominihuman's generated results are characterized by very small motion amplitudes, often closely resembling the reference image's static pose. Our model, conversely, is capable of generating a significantly wider range of motion, thus offering enhanced diversity in output.

\textbf{Consistency of Long Term Generation}

Compared to previous methods that typically generate short, isolated video clips focused on solo speaking scenarios, film-grade video generation demands long-term consistency across multiple generated clips, e.g motion, camera movement and identity preservation. 
Our method utilizes FramePack to encode more motion frames, enabling the model to capture long-term temporal dependencies and, intuitively, achieve better preservation of coherent temporal information.

\begin{figure*}[htbp]
\centering
\includegraphics[width=\textwidth]{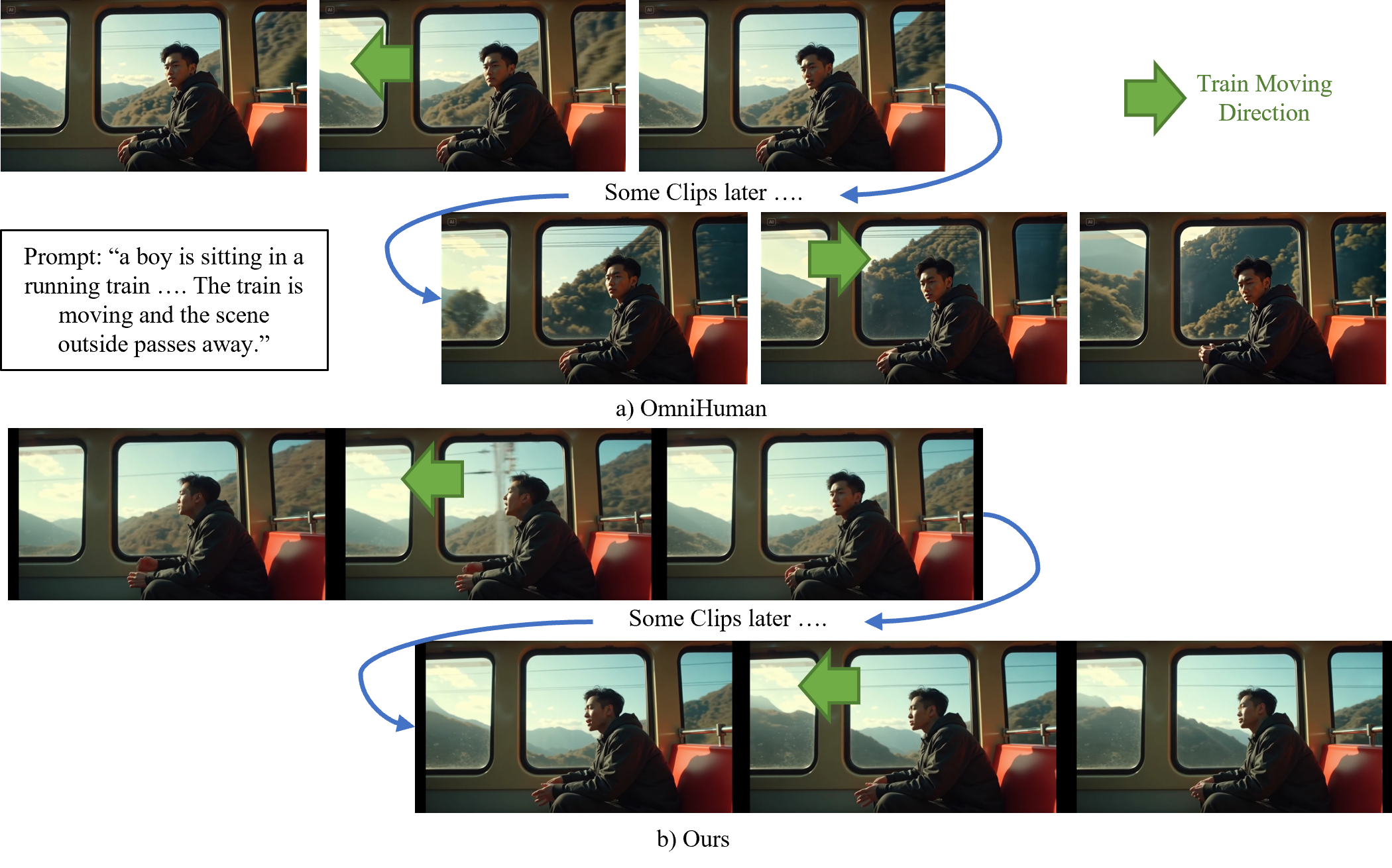}
\caption{
Qualitative comparison of motion preservation performance between our method and OmniHuman.}
\label{fig:long_term}
\end{figure*}

As shown in ~\ref{fig:long_term}, when generating a scene in which the target is required to maintain consistent motion (e.g., a train moving in a coherent direction), OmniHuman fails to preserve the motion trend across multiple clips, while our method successfully maintains consistency in both the direction and speed of the train.

\begin{figure*}[htbp]
\centering
\includegraphics[width=\textwidth]{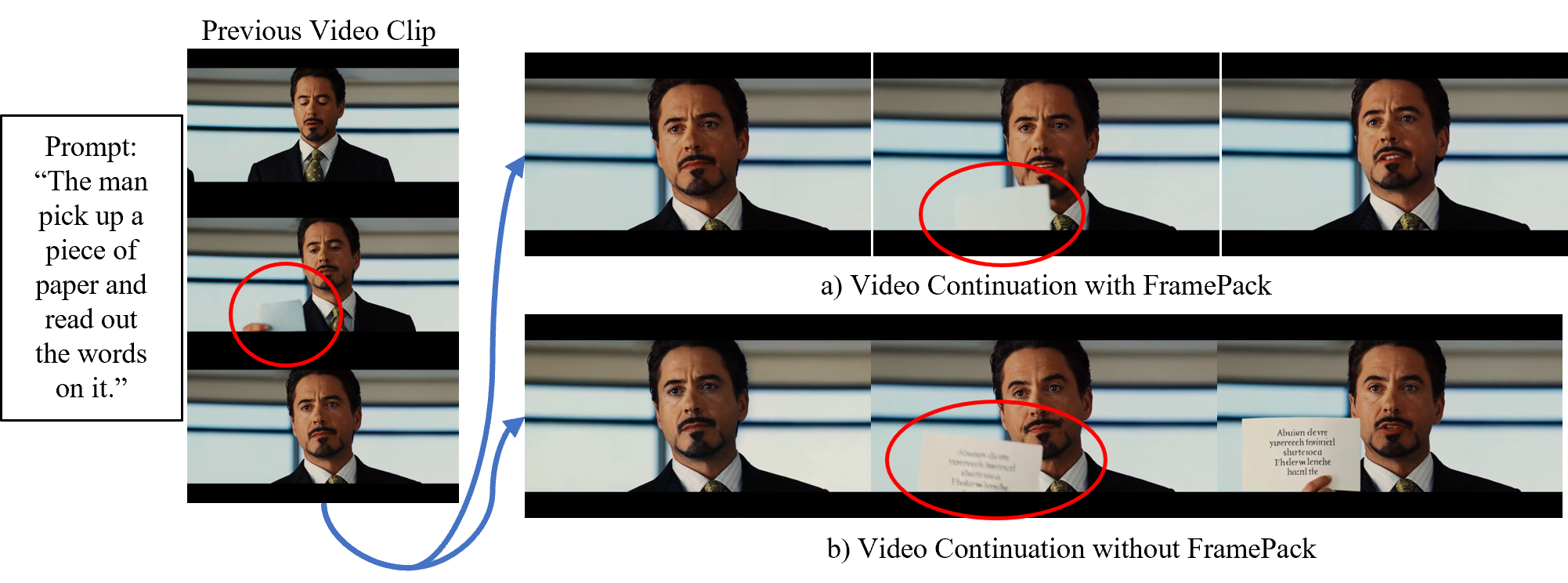}
\caption{
Maintaining item identity across consecutive video clips.}
\label{fig:long_term2}
\end{figure*}

When continuing to generate a new video clip following previously generated ones, the prior clips are used as motion frames. By utilizing FramePack to encode a larger number of motion frames, our method not only preserves the overall motion trend but also helps maintain element identity across clips. For instance, as shown in ~\ref{fig:long_term2}, the generated character picks up a piece of paper that visually matches the one from the previous clip. In contrast, without FramePack, the appearance of the same object may drift significantly.

\subsection{Quantitative Evaluation}

We conduct quantitative comparisons on the EMTD dataset proposed by \cite{echomimicv2},which primarily consists of solo-talking videos, evaluating several open-source audio-animation methods. This includes EchoMimicV2, developed by~\cite{echomimicv2}, and MimicMotion from \cite{mimicmotion2024}. Both of these approaches rely on pre-extracted pose sequences to animate images. Additionally, we compare our work with EMO2, introduced by \cite{emo2}, which employs a two-stage process: generating partial hand motion from audio and subsequently animating the character using both the audio and the generated motion. We also include recent audio-driven DIT-based methods in our comparisons, such as FantasyTalking \cite{Wang2025FantasyTalkingRT} and Hunyuan-Avatar.

To demonstrate the superiority of our proposed method, we evaluate the models using several metrics. We employ Fréchet Inception Distance (FID)~\cite{fid}, SSIM~\cite{ssim}, and PSNR~\cite{psnr} to assess the quality of the generated frames. Fréchet Video Distance (FVD)~\cite{fvd} is used to gauge the overall coherence of the generated videos. To evaluate identity consistency, we calculate the cosine similarity (CSIM) between the facial features of the reference image and the generated video frames. We also utilize Sync-C, as proposed by~\cite{syncnet}, to assess the synchronization quality between lip movements and audio signals. Furthermore, we measure Hand Keypoint Confidence (HKC) to evaluate the quality of hand representation in generated frames, while Hand Keypoint Variance (HKV) serves as an indicator of the richness of hand motion. Additionally, EFID proposed by~\cite{tian2025emo} is adopted to quantitatively assess the divergence in expressions between the synthesized videos and those in the ground truth dataset.

As illustrated in~\ref{tab:quantitative_comp}, our method surpasses the others in terms of frame quality, as indicated by improved image metrics (FID, SSIM, PSNR). Additionally, it demonstrates a clear advantage in video quality assessment, with a lower FVD score. In terms of detail generation, our approach produces clearer and more accurate hand shapes, as reflected by the higher HKC score. Furthermore, it generates more vivid and diverse hand motions, indicated by a higher HKV value.
It is worth noting that EMO2 achieves highest HKC and HKV scores. This can be attributed to the fact that EMO2 generates frames conditioned on pre-generated motion sequences, allowing for better control over hand motion diversity. Moreover, the use of MANO contributes to its superior performance in HKC compared to other methods. On the other hand, HY-Avatar tends to produce characters with "poker-face" expressions, which results in a higher EFID compared to other methods.


\begin{table*}[tb]
  \centering
  \caption{Quantitative comparisons with SOTA.}
  \label{tab:quantitative_comp}
  \setlength{\tabcolsep}{2pt}
  \begin{tabularx}{\textwidth}{l>{\centering\arraybackslash}X>{\centering\arraybackslash}X>{\centering\arraybackslash}X>{\centering\arraybackslash}X>{\centering\arraybackslash}X>{\centering\arraybackslash}X>{\centering\arraybackslash}X>{\centering\arraybackslash}X>{\centering\arraybackslash}X}
    \toprule
    Method & FID$\downarrow$ & FVD$\downarrow$ & SSIM$\uparrow$ & PSNR$\uparrow$ & \small{Sync-C}$\uparrow$ & EFID$\downarrow$ & HKC$\uparrow$ & HKV$\uparrow$ & CSIM$\uparrow$ \\
    \midrule
    EchoMimicV2 & 33.42 & 217.71 & 0.662 & 18.17 &  4.44 &  1.052 & 0.425 & 0.150 & 0.519 \\
    MimicMotion& 25.38 & 248.95 & 0.585 & 17.15 &  2.68 &  0.617 & 0.356 & 0.169 & 0.608 \\
    EMO2 & 27.28 & {129.41} & 0.662 & 17.75 &  4.58 &  {0.218} & 0.553 & {0.198} & {0.650} \\
    FantasyTalking & 22.60 & 178.12 & \text{0.703} & 19.63 & 3.00 &  0.366 & 0.281 & 0.087 & 0.626 \\
    HY-Avatar & 18.07 & 145.77 & 0.670 & 18.16 & {4.71} & 0.7082 & 0.379 & 0.145 & 0.583 \\
    Ours & 15.66 & 129.57 & 0.734 & {20.49} & 4.51 & 0.283 & {0.435} & 0.142 & 0.677 \\
    \bottomrule
  \end{tabularx}
\end{table*}


\section{Conclusion}
This paper presented significant advancements in audio-driven human video generation, specifically addressing the complexities of film and television scenarios. We demonstrated the crucial synergy between text for global motion control and audio for fine-grained character expressions, leading to more expressive and consistent video generation. Our comprehensive approach, from data to training and optimized inference, aims to make high-quality audio-driven video synthesis more accessible and practical. Despite this progress, truly complex film and television challenges, such as nuanced multi-person interactions and precise camera control driven solely by audio, remain formidable. 
Wan-S2V is the first in our Vida research series. We envision this series, including future work on advanced character control and dynamic dancing generation, will foster continued research and development, pushing the boundaries of human-centric video synthesis.

\section{Contributors}
All contributors are listed in alphabetical order by their last names.

\begin{itemize}[nosep,label={},leftmargin=*]
    \bfseries
    \color{blue}
    \item Xin Gao, Li Hu, Siqi Hu, Mingyang Huang, Chaonan Ji, Dechao Meng, Jinwei Qi, Penchong Qiao, Zhen Shen, Yafei Song, Ke Sun, Linrui Tian, Guangyuan Wang, Qi Wang, Zhongjian Wang, Jiayu Xiao, Sheng Xu, Bang Zhang, Peng Zhang, Xindi Zhang, Zhe Zhang, Jingren Zhou, Lian Zhuo
    
\end{itemize}


\FloatBarrier

\bibliography{iclr2021_conference}

\begin{thebibliography}{30}
\providecommand{\natexlab}[1]{#1}
\providecommand{\url}[1]{\texttt{#1}}
\expandafter\ifx\csname urlstyle\endcsname\relax
  \providecommand{\doi}[1]{doi: #1}\else
  \providecommand{\doi}{doi: \begingroup \urlstyle{rm}\Url}\fi

\bibitem[Bai et~al.(2025)Bai, Chen, Liu, Wang, Ge, Song, Dang, Wang, Wang, Tang, Zhong, Zhu, Yang, Li, Wan, Wang, Ding, Fu, Xu, Ye, Zhang, Xie, Cheng, Zhang, Yang, Xu, and Lin]{Qwen2.5-VL}
Shuai Bai, Keqin Chen, Xuejing Liu, Jialin Wang, Wenbin Ge, Sibo Song, Kai Dang, Peng Wang, Shijie Wang, Jun Tang, Humen Zhong, Yuanzhi Zhu, Mingkun Yang, Zhaohai Li, Jianqiang Wan, Pengfei Wang, Wei Ding, Zheren Fu, Yiheng Xu, Jiabo Ye, Xi~Zhang, Tianbao Xie, Zesen Cheng, Hang Zhang, Zhibo Yang, Haiyang Xu, and Junyang Lin.
\newblock Qwen2.5-vl technical report.
\newblock \emph{arXiv preprint arXiv:2502.13923}, 2025.

\bibitem[Chen et~al.(2025)Chen, Liang, Zhou, Huang, Ma, Tang, Lin, Zhou, and Lu]{hu2025HunyuanVideo-Avatar}
Yi~Chen, Sen Liang, Zixiang Zhou, Ziyao Huang, Yifeng Ma, Junshu Tang, Qin Lin, Yuan Zhou, and Qinglin Lu.
\newblock Hunyuanvideo-avatar: High-fidelity audio-driven human animation for multiple characters, 2025.
\newblock URL \url{https://arxiv.org/pdf/2505.20156}.

\bibitem[Chung \& Zisserman(2017)Chung and Zisserman]{syncnet}
Joon~Son Chung and Andrew Zisserman.
\newblock Out of time: automated lip sync in the wild.
\newblock In \emph{Computer Vision--ACCV 2016 Workshops: ACCV 2016 International Workshops, Taipei, Taiwan, November 20-24, 2016, Revised Selected Papers, Part II 13}, pp.\  251--263. Springer, 2017.

\bibitem[Fang \& Zhao(2024)Fang and Zhao]{usp}
Jiarui Fang and Shangchun Zhao.
\newblock Usp: A unified sequence parallelism approach for long context generative ai, 2024.
\newblock URL \url{https://arxiv.org/abs/2405.07719}.

\bibitem[Heusel et~al.(2017)Heusel, Ramsauer, Unterthiner, Nessler, and Hochreiter]{fid}
Martin Heusel, Hubert Ramsauer, Thomas Unterthiner, Bernhard Nessler, and Sepp Hochreiter.
\newblock Gans trained by a two time-scale update rule converge to a local nash equilibrium.
\newblock \emph{Advances in neural information processing systems}, 30, 2017.

\bibitem[Ho et~al.(2020)Ho, Jain, and Abbeel]{ddpm}
Jonathan Ho, Ajay Jain, and Pieter Abbeel.
\newblock Denoising diffusion probabilistic models.
\newblock \emph{Advances in neural information processing systems}, 33:\penalty0 6840--6851, 2020.

\bibitem[Horé \& Ziou(2010)Horé and Ziou]{psnr}
Alain Horé and Djemel Ziou.
\newblock Image quality metrics: Psnr vs. ssim.
\newblock In \emph{2010 20th International Conference on Pattern Recognition}, pp.\  2366--2369, 2010.
\newblock \doi{10.1109/ICPR.2010.579}.

\bibitem[Kong et~al.(2025)Kong, Tian, Zhang, Min, Dai, Zhou, Xiong, Li, Wu, Zhang, Wu, Lin, Yuan, Long, Wang, Wang, Li, Huang, Yang, Tan, Wang, Song, Bai, Wu, Xue, Wang, Wang, Liu, Li, Li, Wang, Yu, Deng, Li, Chen, Cui, Peng, Yu, He, Xu, Zhou, Xu, Tao, Lu, Liu, Zhou, Wang, Yang, Wang, Liu, Jiang, and Zhong]{hunyuanvideo}
Weijie Kong, Qi~Tian, Zijian Zhang, Rox Min, Zuozhuo Dai, Jin Zhou, Jiangfeng Xiong, Xin Li, Bo~Wu, Jianwei Zhang, Kathrina Wu, Qin Lin, Junkun Yuan, Yanxin Long, Aladdin Wang, Andong Wang, Changlin Li, Duojun Huang, Fang Yang, Hao Tan, Hongmei Wang, Jacob Song, Jiawang Bai, Jianbing Wu, Jinbao Xue, Joey Wang, Kai Wang, Mengyang Liu, Pengyu Li, Shuai Li, Weiyan Wang, Wenqing Yu, Xinchi Deng, Yang Li, Yi~Chen, Yutao Cui, Yuanbo Peng, Zhentao Yu, Zhiyu He, Zhiyong Xu, Zixiang Zhou, Zunnan Xu, Yangyu Tao, Qinglin Lu, Songtao Liu, Daquan Zhou, Hongfa Wang, Yong Yang, Di~Wang, Yuhong Liu, Jie Jiang, and Caesar Zhong.
\newblock Hunyuanvideo: A systematic framework for large video generative models, 2025.
\newblock URL \url{https://arxiv.org/abs/2412.03603}.

\bibitem[Li et~al.(2024)Li, Xu, Zhan, Mu, Li, Cheng, Chen, Chen, Ye, Wang, and Zhu]{OpenHumanVid}
Hui Li, Mingwang Xu, Yun Zhan, Shan Mu, Jiaye Li, Kaihui Cheng, Yuxuan Chen, Tan Chen, Mao Ye, Jingdong Wang, and Siyu Zhu.
\newblock Openhumanvid: A large-scale high-quality dataset for enhancing human-centric video generation, 2024.

\bibitem[Liao et~al.(2023)Liao, Duan, Feng, Zhao, Yang, and Chen]{light-asd}
Junhua Liao, Haihan Duan, Kanghui Feng, Wanbing Zhao, Yanbing Yang, and Liangyin Chen.
\newblock A light weight model for active speaker detection.
\newblock In \emph{Proceedings of the IEEE/CVF Conference on Computer Vision and Pattern Recognition (CVPR)}, pp.\  22932--22941, June 2023.

\bibitem[Lin et~al.(2025)Lin, Jiang, Yang, Zheng, and Liang]{ominihuman}
Gaojie Lin, Jianwen Jiang, Jiaqi Yang, Zerong Zheng, and Chao Liang.
\newblock Omnihuman-1: Rethinking the scaling-up of one-stage conditioned human animation models, 2025.
\newblock URL \url{https://arxiv.org/abs/2502.01061}.

\bibitem[Lipman et~al.(2023)Lipman, Chen, Ben-Hamu, Nickel, and Le]{lipman2023flowmatchinggenerativemodeling}
Yaron Lipman, Ricky T.~Q. Chen, Heli Ben-Hamu, Maximilian Nickel, and Matt Le.
\newblock Flow matching for generative modeling, 2023.
\newblock URL \url{https://arxiv.org/abs/2210.02747}.

\bibitem[Meng et~al.(2024)Meng, Zhang, Li, and Ma]{echomimicv2}
Rang Meng, Xingyu Zhang, Yuming Li, and Chenguang Ma.
\newblock Echomimicv2: Towards striking, simplified, and semi-body human animation, 2024.
\newblock URL \url{https://arxiv.org/abs/2411.10061}.

\bibitem[Schneider et~al.(2019)Schneider, Baevski, Collobert, and Auli]{wav2vec}
Steffen Schneider, Alexei Baevski, Ronan Collobert, and Michael Auli.
\newblock wav2vec: Unsupervised pre-training for speech recognition.
\newblock pp.\  3465--3469, 09 2019.
\newblock \doi{10.21437/Interspeech.2019-1873}.

\bibitem[Schuhmann(2022)]{improved-aesthetic-predictor}
Christoph Schuhmann.
\newblock improved-aesthetic-predictor.
\newblock \url{https://github.com/christophschuhmann/improved-aesthetic-predictor}, 2022.

\bibitem[Tian et~al.(2024)Tian, Wang, Zhang, and Bo]{tian2024emo}
Linrui Tian, Qi~Wang, Bang Zhang, and Liefeng Bo.
\newblock Emo: Emote portrait alive - generating expressive portrait videos with audio2video diffusion model under weak conditions, 2024.

\bibitem[Tian et~al.(2025{\natexlab{a}})Tian, Hu, Wang, Zhang, and Bo]{emo2}
Linrui Tian, Siqi Hu, Qi~Wang, Bang Zhang, and Liefeng Bo.
\newblock Emo2: End-effector guided audio-driven avatar video generation, 2025{\natexlab{a}}.
\newblock URL \url{https://arxiv.org/abs/2501.10687}.

\bibitem[Tian et~al.(2025{\natexlab{b}})Tian, Wang, Zhang, and Bo]{tian2025emo}
Linrui Tian, Qi~Wang, Bang Zhang, and Liefeng Bo.
\newblock Emo: Emote portrait alive generating expressive portrait videos with audio2video diffusion model under weak conditions.
\newblock In \emph{European Conference on Computer Vision}, pp.\  244--260. Springer, 2025{\natexlab{b}}.

\bibitem[Unterthiner et~al.(2019)Unterthiner, van Steenkiste, Kurach, Marinier, Michalski, and Gelly]{fvd}
Thomas Unterthiner, Sjoerd van Steenkiste, Karol Kurach, Rapha{\"e}l Marinier, Marcin Michalski, and Sylvain Gelly.
\newblock Fvd: A new metric for video generation.
\newblock 2019.

\bibitem[Wan et~al.(2025)Wan, Wang, Ai, Wen, Mao, Xie, Chen, Yu, Zhao, Yang, Zeng, Wang, Zhang, Zhou, Wang, Chen, Zhu, Zhao, Yan, Huang, Feng, Zhang, Li, Wu, Chu, Feng, Zhang, Sun, Fang, Wang, Gui, Weng, Shen, Lin, Wang, Wang, Zhou, Wang, Shen, Yu, Shi, Huang, Xu, Kou, Lv, Li, Liu, Wang, Zhang, Huang, Li, Wu, Liu, Pan, Zheng, Hong, Shi, Feng, Jiang, Han, Wu, and Liu]{wan2025}
Team Wan, Ang Wang, Baole Ai, Bin Wen, Chaojie Mao, Chen-Wei Xie, Di~Chen, Feiwu Yu, Haiming Zhao, Jianxiao Yang, Jianyuan Zeng, Jiayu Wang, Jingfeng Zhang, Jingren Zhou, Jinkai Wang, Jixuan Chen, Kai Zhu, Kang Zhao, Keyu Yan, Lianghua Huang, Mengyang Feng, Ningyi Zhang, Pandeng Li, Pingyu Wu, Ruihang Chu, Ruili Feng, Shiwei Zhang, Siyang Sun, Tao Fang, Tianxing Wang, Tianyi Gui, Tingyu Weng, Tong Shen, Wei Lin, Wei Wang, Wei Wang, Wenmeng Zhou, Wente Wang, Wenting Shen, Wenyuan Yu, Xianzhong Shi, Xiaoming Huang, Xin Xu, Yan Kou, Yangyu Lv, Yifei Li, Yijing Liu, Yiming Wang, Yingya Zhang, Yitong Huang, Yong Li, You Wu, Yu~Liu, Yulin Pan, Yun Zheng, Yuntao Hong, Yupeng Shi, Yutong Feng, Zeyinzi Jiang, Zhen Han, Zhi-Fan Wu, and Ziyu Liu.
\newblock Wan: Open and advanced large-scale video generative models.
\newblock \emph{arXiv preprint arXiv:2503.20314}, 2025.

\bibitem[Wang et~al.(2025)Wang, Wang, Jiang, Fan, Zhang, Qi, Zhao, and Xu]{Wang2025FantasyTalkingRT}
Mengchao Wang, Qiang Wang, Fan Jiang, Yaqi Fan, Yunpeng Zhang, Yonggang Qi, Kun Zhao, and Mu~Xu.
\newblock Fantasytalking: Realistic talking portrait generation via coherent motion synthesis.
\newblock \emph{ArXiv}, abs/2504.04842, 2025.
\newblock URL \url{https://api.semanticscholar.org/CorpusID:277621659}.

\bibitem[Wang et~al.(2024)Wang, Shi, Ou, Chen, Lin, Wang, Jiang, Yang, Zheng, Tao, Yang, Wan, and Zhang]{koala36m}
Qiuheng Wang, Yukai Shi, Jiarong Ou, Rui Chen, Ke~Lin, Jiahao Wang, Boyuan Jiang, Haotian Yang, Mingwu Zheng, Xin Tao, Fei Yang, Pengfei Wan, and Di~Zhang.
\newblock Koala-36m: A large-scale video dataset improving consistency between fine-grained conditions and video content, 2024.
\newblock URL \url{https://arxiv.org/abs/2410.08260}.

\bibitem[Wang et~al.(2004)Wang, Bovik, Sheikh, and Simoncelli]{ssim}
Zhou Wang, A.C. Bovik, H.R. Sheikh, and E.P. Simoncelli.
\newblock Image quality assessment: from error visibility to structural similarity.
\newblock \emph{IEEE Transactions on Image Processing}, 13\penalty0 (4):\penalty0 600--612, 2004.
\newblock \doi{10.1109/TIP.2003.819861}.

\bibitem[Wu et~al.(2023)Wu, Zhang, Liao, Chen, Hou, Wang, Sun, Yan, and Lin]{dover}
Haoning Wu, Erli Zhang, Liang Liao, Chaofeng Chen, Jingwen~Hou Hou, Annan Wang, Wenxiu~Sun Sun, Qiong Yan, and Weisi Lin.
\newblock Exploring video quality assessment on user generated contents from aesthetic and technical perspectives.
\newblock In \emph{International Conference on Computer Vision (ICCV)}, 2023.

\bibitem[Xu et~al.(2023)Xu, Zhang, Cai, Rezatofighi, Yu, Tao, and Geiger]{unimatch}
Haofei Xu, Jing Zhang, Jianfei Cai, Hamid Rezatofighi, Fisher Yu, Dacheng Tao, and Andreas Geiger.
\newblock Unifying flow, stereo and depth estimation.
\newblock \emph{IEEE Transactions on Pattern Analysis and Machine Intelligence}, 2023.

\bibitem[Xu et~al.(2022)Xu, Zhang, Zhang, and Tao]{vitpose}
Yufei Xu, Jing Zhang, Qiming Zhang, and Dacheng Tao.
\newblock Vitpose: Simple vision transformer baselines for human pose estimation, 2022.
\newblock URL \url{https://arxiv.org/abs/2204.12484}.

\bibitem[Yang et~al.(2023)Yang, Zeng, Yuan, and Li]{dwpose}
Zhendong Yang, Ailing Zeng, Chun Yuan, and Yu~Li.
\newblock Effective whole-body pose estimation with two-stages distillation.
\newblock In \emph{Proceedings of the IEEE/CVF International Conference on Computer Vision}, pp.\  4210--4220, 2023.

\bibitem[Zhang \& Agrawala(2025)Zhang and Agrawala]{zhang2025framepack}
Lvmin Zhang and Maneesh Agrawala.
\newblock Packing input frame contexts in next-frame prediction models for video generation.
\newblock \emph{Arxiv}, 2025.

\bibitem[Zhang et~al.(2024)Zhang, Gu, Wang, Wang, Cheng, Zhu, and Zou]{mimicmotion2024}
Yuang Zhang, Jiaxi Gu, Li-Wen Wang, Han Wang, Junqi Cheng, Yuefeng Zhu, and Fangyuan Zou.
\newblock Mimicmotion: High-quality human motion video generation with confidence-aware pose guidance.
\newblock \emph{arXiv preprint arXiv:2406.19680}, 2024.

\bibitem[Zhao et~al.(2023)Zhao, Gu, Varma, Luo, Huang, Xu, Wright, Shojanazeri, Ott, Shleifer, Desmaison, Balioglu, Damania, Nguyen, Chauhan, Hao, Mathews, and Li]{fsdp}
Yanli Zhao, Andrew Gu, Rohan Varma, Liang Luo, Chien-Chin Huang, Min Xu, Less Wright, Hamid Shojanazeri, Myle Ott, Sam Shleifer, Alban Desmaison, Can Balioglu, Pritam Damania, Bernard Nguyen, Geeta Chauhan, Yuchen Hao, Ajit Mathews, and Shen Li.
\newblock Pytorch fsdp: Experiences on scaling fully sharded data parallel, 2023.
\newblock URL \url{https://arxiv.org/abs/2304.11277}.

\end{thebibliography}
\bibliographystyle{iclr2021_conference}


\end{document}